\documentclass[letterpaper, 10 pt, conference]{ieeeconf}
\IEEEoverridecommandlockouts                              
                                                          
\usepackage[noadjust]{cite}

\usepackage{subfig}
\usepackage{amsmath,amssymb,amsfonts}
\usepackage{graphicx}
\usepackage[noend]{algpseudocode}
\usepackage[table]{xcolor}

\usepackage{caption}

\DeclareCaptionFormat{algor}{%
  \hrulefill\par\offinterlineskip\vskip1pt%
    \textbf{#1#2}#3\offinterlineskip\hrulefill}
\DeclareCaptionStyle{algori}{singlelinecheck=off,format=algor,labelsep=space}

\title{\LARGE \bf Robustness for Free: Quality-Diversity Driven Discovery of Agile Soft Robotic Gaits}
\author{John Daly$^{1}$,  Daniel Casper$^{1}$,  Muhammad Farooq$^{1}$, Andrew James$^{1}$, Ali Khan$^{1}$, \\ Phoenix Mulgrew$^{1}$, Daniel Tyebkhan$^{1}$,  Bao Vo$^{1}$, John Rieffel$^{2}$
\thanks{$^{1}$Undergraduate Student,  Union College, Schenectady, NY }%
\thanks{$^{2}$  Professor, Computer Science Department, Union College, Schenectady, NY
        {\tt\small rieffelj@union.edu}}%
}

\begin{document}

\maketitle
\thispagestyle{empty}
\pagestyle{empty}

\begin{abstract}
Soft robotics aims to develop robots able to adapt their behavior across a wide range of unstructured and unknown environments.   A critical challenge of soft robotic control is that  nonlinear dynamics often result in complex behaviors hard to model and predict.  Typically behaviors for mobile soft robots are discovered through empirical trial and error and hand-tuning.  More recently, optimization algorithms such as Genetic Algorithms (GA) have been used to discover gaits, but  these behaviors are often optimized for a single environment or terrain, and can be brittle to unplanned changes to terrain.  In this paper we demonstrate how Quality Diversity Algorithms, which search of a range of high-performing behaviors, can produce repertoires of gaits that are robust to changing terrains.  This robustness significantly out-performs that of gaits produced by a single objective optimization algorithm.
\end{abstract}


\begin{keywords}
Soft Robotics, Quality Diversity Algorithms, Evolutionary Robotics
\end{keywords}

%
\IEEEpeerreviewmaketitle

\section{Introduction}

The field of soft robotics seeks to create bio-inspired soft systems whose performance rivals that of natural systems.  Such robots hold a particular appeal for tasks in unstructured and uncertain environments, where compliance and deformability might allow a robot to change shape to squeeze through small apertures, or survive large drops.   However, because they are nature high dimensional dynamic systems with an essentially infinite number of degrees of freedom, they are difficult to control by conventional means\cite{lipson2014challenges,shepherd2011multigait}.

As a result of this complexity, the gaits of mobile soft robots are typically discovered through empirical trial-and-error\cite{shepherd2011multigait}, a process that can be both challenging and time consuming.  More recently, optimization algorithms such have Genetic Algorithms have been used to automatically generate gaits for soft robots \cite{rieffel2009evolving, rieffel2014growing, cochevelou2023differentiable, bhatia2021evolution}  but like the hand-tuned ones they tend to be optimized for only a single terrain, and can therefore be brittle to changes.  This brittleness is a challenge to creating soft robots able to autonomously adapt their behavior as needed, for instance when they encounter a novel environmental substrate, or when they are physically damaged.  

One promising alternative approach is to use novelty-oriented approaches such as Quality Diversity Algorithms (QDAs) to develop repertoires of effective but behaviorally distinct gaits.  One advantage of these approaches might be that the behavioral diversity of the repertoire offers some insurance to changes in terrain. In this paper we demonstrate the non-robust nature of solutions evolved with a state-of-the-art optimization algorithm (CMA-ES), and explore how Quality Diversity Algorithms can not just discover effective gaits, but can also be robust to significant changes in their environment.  

\section{Background and Related Work}

\subsection{Soft Robotics Simulators}

Although the complexity of soft materials makes soft robots incredibly difficult to simulate with fidelity sufficient for ``sim-to-real transfer''~\cite{kriegman2020scalable}  one can still glean significant insights about soft robotics by generating gaits in simulation.  Several soft robotic simulators exist, including VoxCraft~\cite{liu_voxcraft_2020,kriegman2020scalable}, SOFA~\cite{duriez2013control,navarro2020model}, EvoGym~\cite{bhatia2021evolution} and more recently differentiable Mass Particle Method (MPM) simulators such as ChainQueen~\cite{hu2019chainqueen} and DiffTaiChi~\cite{hu2019difftaichi}. 

For this work we rely upon {\bf 2D-VSR-Sim}, a 2D Java-based simulator that models a single voxel as a deformable square with four rigid  masses connected by both spring–damper systems (SDSs) and tensile cords~\cite{medvet20202d,medvet2020design}.  Robot morphologies in the simulator are constructed by connecting multiple voxels together at the edges.   The coefficients (mass, spring constants, etc) can be independently configured for each voxel, allowing for a significant range of soft material properties.  Robots in 2D-VSR-Sim are typically controlled by applying oscillating volumetric changes to individual voxels.  These oscillations can controlled in either an open-loop or closed-loop fashion during simulation.

\subsection{Optimization Methods}

Because of the complexity of soft materials and the non-intuitive nature of their behavior, optimization techniques such as Genetic Algorithms (GAs) are a promising approach for the automated discovery of soft robotic gaits.  One advantage is that they can reduce the bias inherent in the human-oriented empirical trial-and-error that is often used in soft robotics.  GAs have been successfully used to evolve soft robotic gaits~\cite{rieffel2009evolving, rieffel2014growing, cochevelou2023differentiable, bhatia2021evolution}.  Because they often require thousands (or more) evaluations to discover effective behaviors, GAs and other optimizers often rely simulators to produce results.

For this work we consider two types of optimization method:

\paragraph{CMA-ES}:  Covariance Matrix Adaptation Evolution Strategies (CMA-ES) is a well established off-the-shelf single-objective optimizer.  We use it in our work as a baseline optimizer against which to compare our QDA approach.

\paragraph{Quality Diversity Algorithms}

Unlike single-objective optimizers like CMA-ES, Quality Diversity algorithms (QDAs)~\cite{stanley-qd,duarte2018evolution,lehman2011evolving,cully2015robots} search for a combination of both {\em novelty} and {\em quality}.  When applied to robotics, this means finding mappings between a robot's {\em parameter space} (those features that can be affected by a controller), and its {\em behavior space} (a description of the outcomes of those actions).
The outcome of a quality diversity search is an {\em archive} describing the variety of behaviors that the robot is capable of -- essentially a mapping between parameter space and behavior space.

QD algorithms have been employed with considerable success with both simulated virtual agents~\cite{lehman2011evolving, duarte2018evolution} and physical robots~\cite{cully2015robots}.  Most notably, Cully~{\em et al.}~\cite{cully2015robots} used a hybrid simulation/physical approach to behavior build repertoires for a hexapod robot.  

Recently, in our own work ~\cite{doney2020behavioral} we explored methods to {\em autonomously} discover novel and effective soft robotic locomotive behaviors by using QDAs.  Specifically, we demonstrated the ability to to autonomously discover a diverse {\em repertoire} of unique locomotive behaviors for a tether-free soft tensegrity robot.  The behavior space in question described the translational and rotational displacement of the robot along the 2D plane.    These discovered behaviors could, in turn, be incorporated into higher-level control strategies such as policy networks~\cite{duarte2018evolution}.

Our interest in this work is not about discovering a repertoire of behaviors producing different displacements, but instead to explore the variety of qualitatively different and interesting gaits a 2D voxel-based robot can traverse a horizontal landscape.

\section{Methods} 

Our ambition was to explore the {\em robustness} of gaits optimized via a conventional single-objective optimizer (CMA-ES) against the archives of behaviors produced by a Quality Diversity Algorithm, specifically the popular QDpy package\cite{qdpy}. Robustness in this case means the loss (or gain) of fitness when a gait evolved on one terrain is transferred to a novel terrain without any subequent re-training.

To explore this, we ran each optimization method on a fixed robot morphology across a variety of terrains, and explored how each optimized system responded to subsequent post-optimization changes in terrain.  In each case, the optimization method was given a fixed evaluation budget in order to make meaningful comparisons between the two approaches.

\subsection{Designing Terrains}

\begin{figure*}[t!]
\centering
\subfloat[Flat]{\includegraphics[width=2.2in]{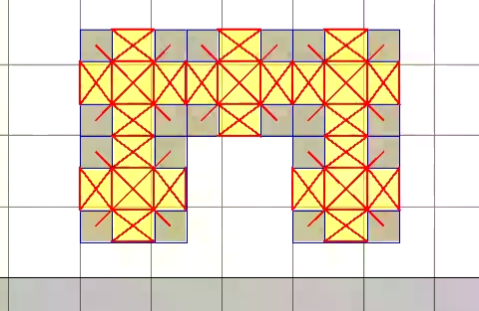} %
}
\hfil
\subfloat[Spiky]{\includegraphics[width=2.2in]{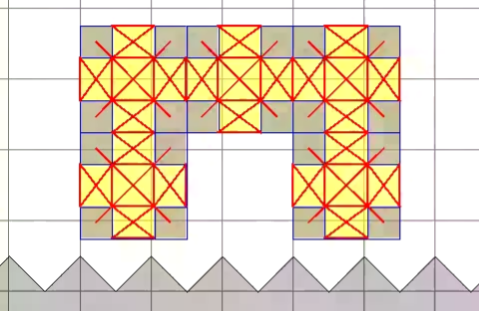}%
}
\hfil
\subfloat[Longspikes]{\includegraphics[width=2.2in]{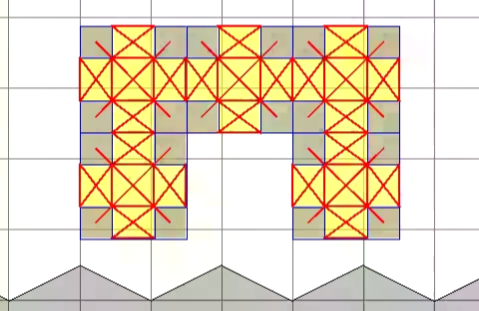}
}
\\
\subfloat[Longerspikes]{\includegraphics[width=2.2in]{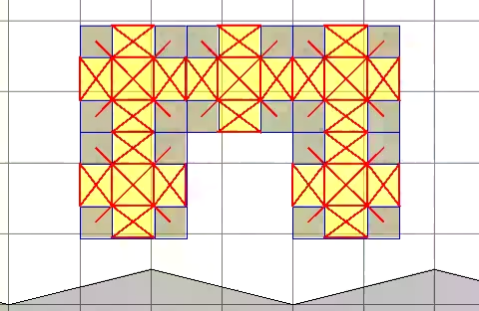}
}
\hfil
\subfloat[Sawtooth]{\includegraphics[width=2.2in]{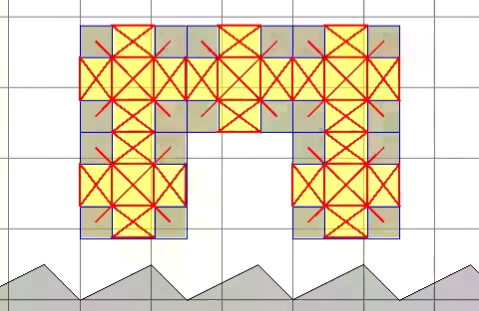}
}
\hfil
\subfloat[Valley]{\includegraphics[width=2.2in]{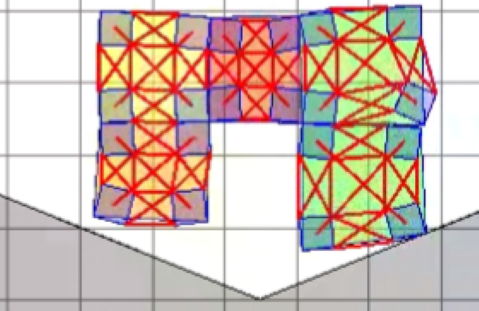}
}
\caption{The six terrains used in this study.}
\label{fig:terrains}
\end{figure*}

The first step for designing this experiment was creating unique terrains for the gaits to be evolved on.  Terrains were generated as periodic waves, and  described in 2D-VSR-Sim as a sequence of coordinate points along the plane.  Terrains were chosen to be unique and variable, but not impossible to traverse.  The surfaces were thefore smaller in dimension than the robot itself.  The six terrains chosen, shown in Figure~\ref{fig:terrains} were:
\begin{itemize}
    \item {\bf flat}: a flat horizontal surface 
    \item {\bf spiky}: periodic triangles 1 unit wide and 0.5 units tall
    \item {\bf longspikes}: periodic triangles 2 units wide and 0.5 units tall. 
    \item {\bf longerspikes}: periodic triangles 4 units wide and 0.5 units tall.  
    \item {\bf sparsespike}: a periodic asymmetric sawtooth shape 1.5  units wide and 0.5 units tall. 
    \item {\bf valley}: an aperiodic valley sloping upwards from the origin in each direction.
\end{itemize}

\subsection{Robot Construction and Actuation}

As described above, 2D-VSR-Sim is a voxel-based simulator, and robot morphologies consist of groupings of connected voxels which can be individually actuated (or passive).  For a given robot design, actuation is achieved by expanding and contracting the actuated voxels.  Details on the simulator are provided in \cite{medvet20202d}.

\begin{figure}[h]
\centering
\includegraphics[width=2.5in]{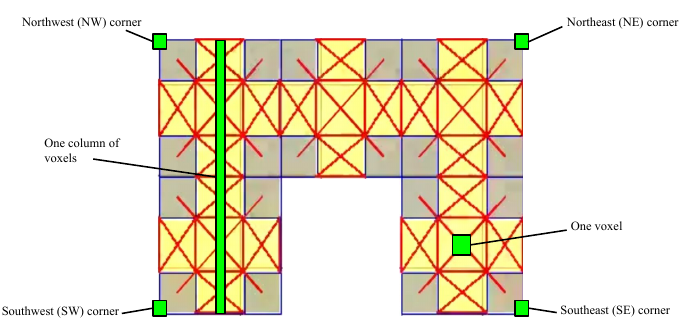}
\caption{The anatomy and shape of the 5-voxel robot used in this work.  Each of the three column of voxels for the robot was independently actuated with a sinusoidal wave.} \label{fig:biped}
\end{figure}

The robots used in this experiment were constructed from 5 voxels in a simple biped-shape, as shown in Figure~\ref{fig:biped}

For this work the robot was actuated by a sinusoidal oscillators controlling each of the three vertical columns of voxels.  The specific actuation parameters varied in order to produce gaits were:

\begin{itemize}
    \item {\bf amplitude:} a global property shared by all voxels  describing how much voxels expand and contract over a given oscillation.
    \item {\bf frequency:} a second global value describing the rate at which all voxels oscillate.
    \item {\bf column phase:} each vertical column of the robot has its own phase offset, causing different columns of the robot to oscillate differently. 
\end{itemize}

These five parameters, encoded as floating point values in the range (0,1), were used both as the {\em genotype} for CMA-ES and the {\em parameter space} for QDPy.

\subsection{Fitness}

The fitness, used for both CMA-ES and QDA was the speed (in units per second) at which the robot traveled horizontally during a gait's simulated evaluation period.   Absolute horizontal speed was used here to avoid biasing solutions towards a ``front'' or ``back'' of the robot --  having a behavior that moves in a subjectively ``backwards'' direction, for instance to the left, is simply a matter of perspective. 

\subsection{Behavioral Descriptors}

The one feature that distinguishes QDAs from optimizers is the use of a behavior-space descriptor, which maps some qualitative measure of {\em how} a gait behaves down into a vector of numerical values.  These descriptor vectors are then used to populate the archive of elite solutions.

\begin{figure}[h]
\centering
\includegraphics[width=0.9\columnwidth]{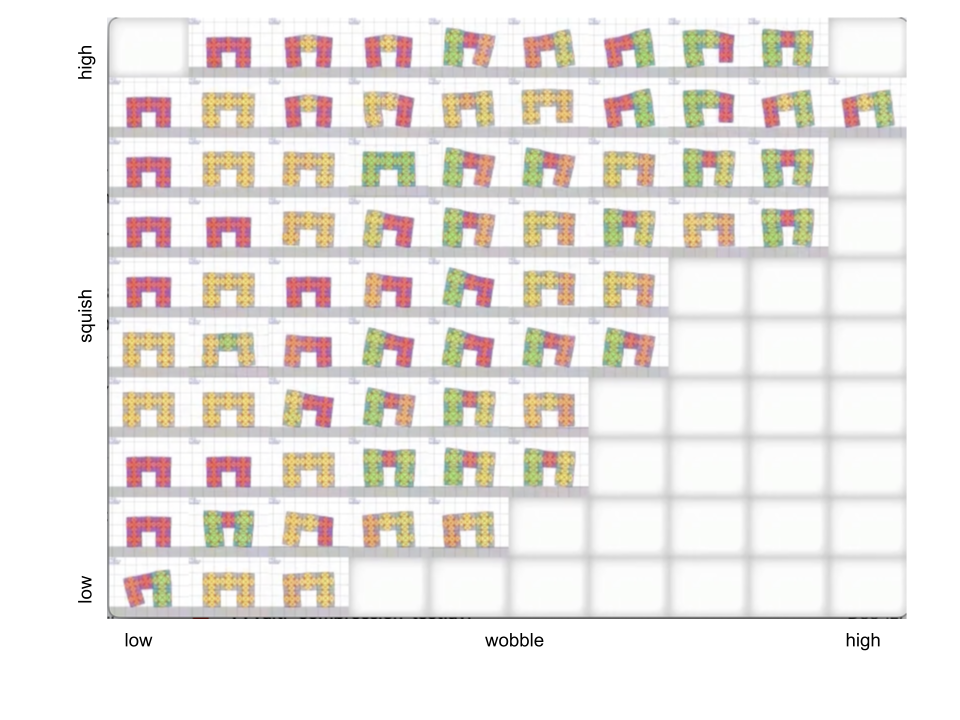}
\caption{The Quality Diversity Algorithm generates an archive of high fitness but behaviorally distinct.  The motion of the robot during the gait is classified along two dimensions - ``squish'' (variance in diagonal width) and ``wobble'' (variation in pitch around the center of mass of the robot), and then placed into a corresponding discreted bin.  Whenever two gaits are placed into the same bin of the archive only the fittest of the two is retained in the archive.}\label{fig:archive}
\end{figure}

In order to facilitate searching for quality alongside diversity though, it's necessary to represent the behavior space as a discrete mapping of ``behavioral niches'' or bins in which all behaviors inside the niche are similar enough that one behavior can reasonably represent the whole subset. The behavior that lies within a niche and is the best performing seen so far is called the {\em elite} of the niche.  The collection of all the elites is known as an {\em archive}.   Our behavior space for this work is divided into 10 regions along each axis, creating 100 distinct niches, as illustrated by Figure~\ref{fig:archive}.

\begin{figure}[h]
\centering
\includegraphics[width=2.0in]{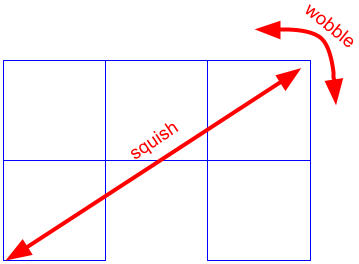}
\caption{Two qualitative measures were used to describe a gait, ``squish'' (variance in distance between two corners), and ``wobble'' (variance in rotation around the center of mass. These descriptors were used to place gaits into the QDA archive.}\label{fig:squish}
\end{figure}

Because our interest was in describing different properties of a soft robotic gait  we chose descriptors that measured various characteristics of soft bodies.  These descriptors, as illustrated in Figure~\ref{fig:squish} are as follows:

\begin{itemize}
    \item {\bf squish}: which measures the variance in the distance between the bottom-left (south-west) and top-right (north-east) corners of the robot.  A gait that exhibits low ``squish'' has low variance in this distance during evaluation, whereas a robot with high ``squish'' varies significantly.
    \item {\bf wobble}: which measures the variance in rotation, or {\em pitch} of the robot relative to the center.  A low-wobble gait remains relatively steady in pitch during evaluation, whereas a high-wobble gait tips back and forth significantly.
\end{itemize}

\subsection{Experimental Procedure}

With the parameters determining actuation established,  as well as a quantitative measure of fitness and qualitative measures of behavior, we can now compare the robustness of gaits evolved under the two treatments: CMA-ES and QDA.  Specifically we are seeking to measure how gaits evolved on one terrain perform when transferred to a novel terrain.

Each algorithm was given the same budget for evolution: 25 seconds of simulation time for each individual evaluation, and a maximum of 2000 evaluations during each optimizing run. Each algorithm was also given the same population/batch size (20).

In the case of CMA-ES this means that the optimizer ran with a population size of 20 and for a maximum of 2000 total evaluations, using horizontal speed as a fitness function.

In the case of the QDA, this means that algorithm ran with a batch size of 20 for a maximum of 2000 total evaluations, using horizontal speed as fitness, and the two behavioral descriptors described above to define the archive.

We then followed these steps to compare the robustness across terrains of the QDA and CMA-ES, as described in the algorithm below.   The process can be broken down into th phases. For an optimization phase, each algorithm was run for its complete evaluation budget on a pre-determined terrain, recording the highest fitness gait (CMA-ES) and highest-fitness gait from the archive (QDA) at the end.  For a transfer phase, the highest-fitness CMA-ES gait was re-evaluated on each of the other terrains, recording the new fitness.   By contrast, the entire archive of elites from the QDA was re-evaluated, and the fitness of the highest-performing elite was recorded.  We repeated this 30 times for each terrain, and repeated the entire process across all six terrains.

\begin{samepage}
\begin{center}

\begin{algorithmic}[h]
\Procedure{Experimental Procedure}{}
\State $T \gets All\:Defined\:Terrains$
\For{$t \in T$}
    \State (Optimization Phase)
    \For{$n = 1, \dots, 30 trials$}
    \State Generate QDA gait archive on terrain $t$
    \State Evolve CMA-ES gait on terrain $t$ 
\EndFor
    \State (Transfer Phase)
    \State $R \gets T - t$
    \For{$r \in R$}
        \State Retest collected QDA archives on $r$
        \State Retest collected CMA-ES behaviors on $r$
    \EndFor
\EndFor
\EndProcedure
\end{algorithmic}
\end{center}
\end{samepage}



All experiments were run on baremetal compute instances provided by the Chameleon Cloud project~\cite{keahey2020lessons}.

With these data gathered we can now compare the fitness gain/loss for each CMA-ES gait before and after transfer against the best new fitness from the corresponding QDA elite archive, averaged across all 30 trials.

\section{Results}

\begin{figure*}[t]
\centering
\includegraphics[width=3.4in]{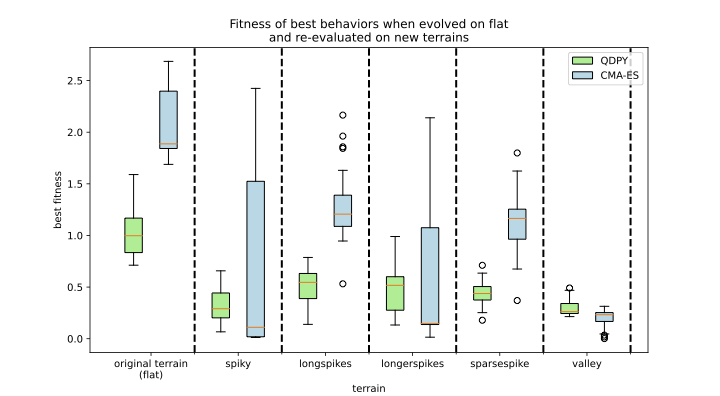}
\includegraphics[width=3.4in]{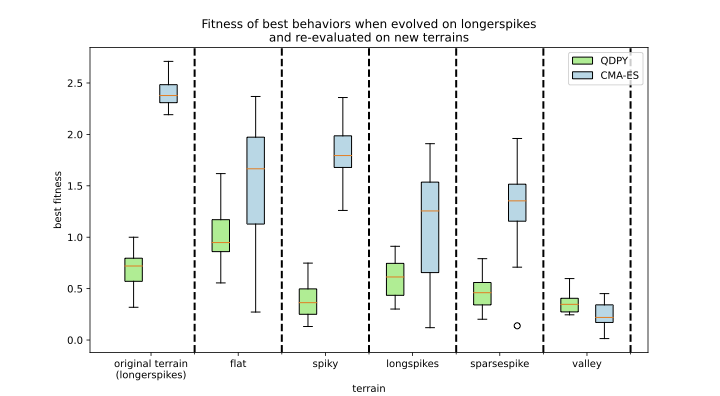} \\
\includegraphics[width=3.4in]{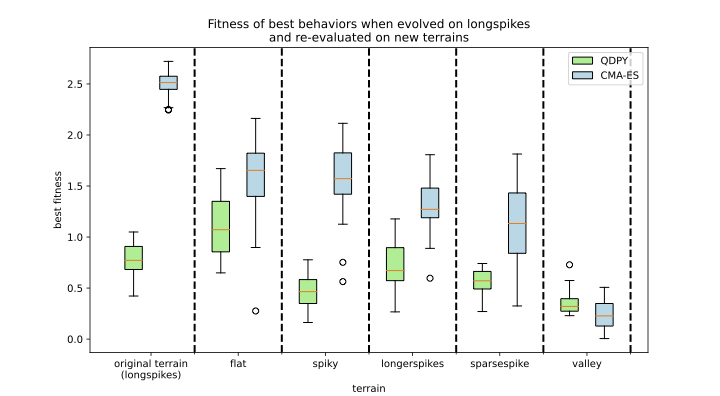} 
\includegraphics[width=3.4in]{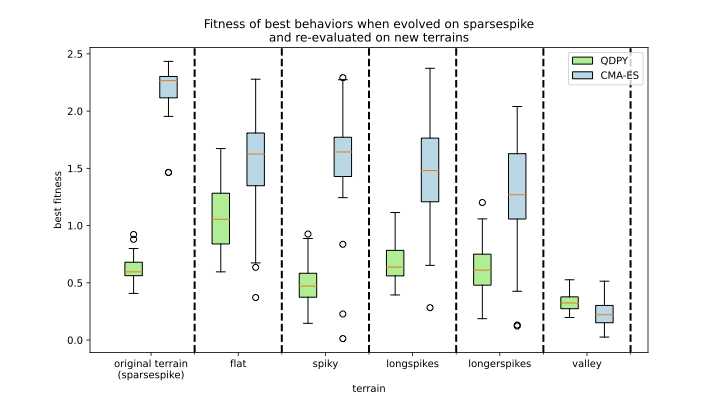}\\
\includegraphics[width=3.4in]{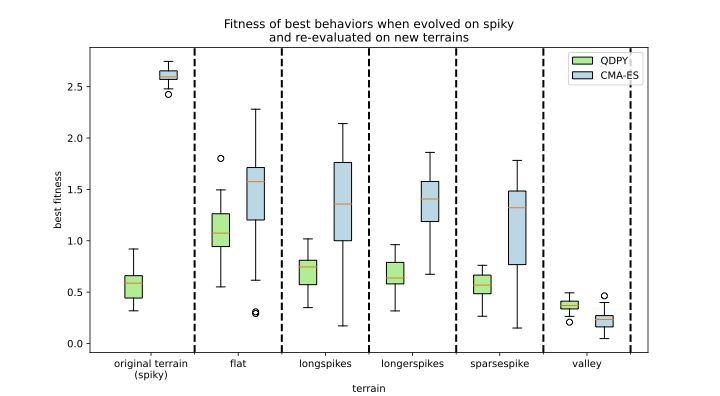}
\includegraphics[width=3.4in]{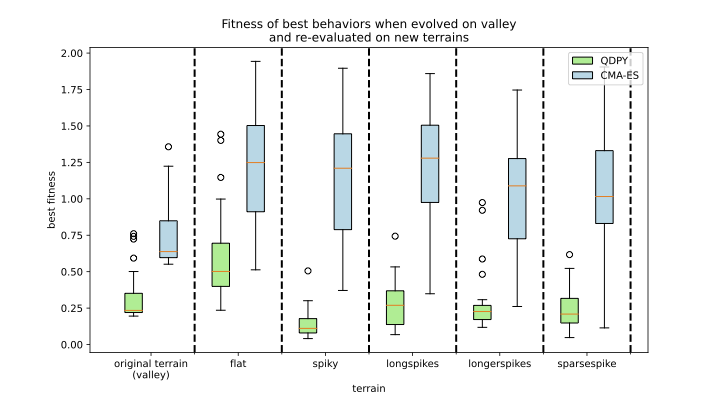}
\caption{Results of generating solutions one one terrain (left-most column in each subfigure and then re-evaluating those solutions on each of the other terrains.  While CMA-ES gaits generally outperforms the QDA, they are quite brittle when transferred to new terrains.  QDA results have lower initial fitness, but generally are much more robust to changes in terrain.}
\label{fig:boxplots}
\end{figure*}

\begin{figure*}[t]
\centering
\includegraphics[width=3.4in]{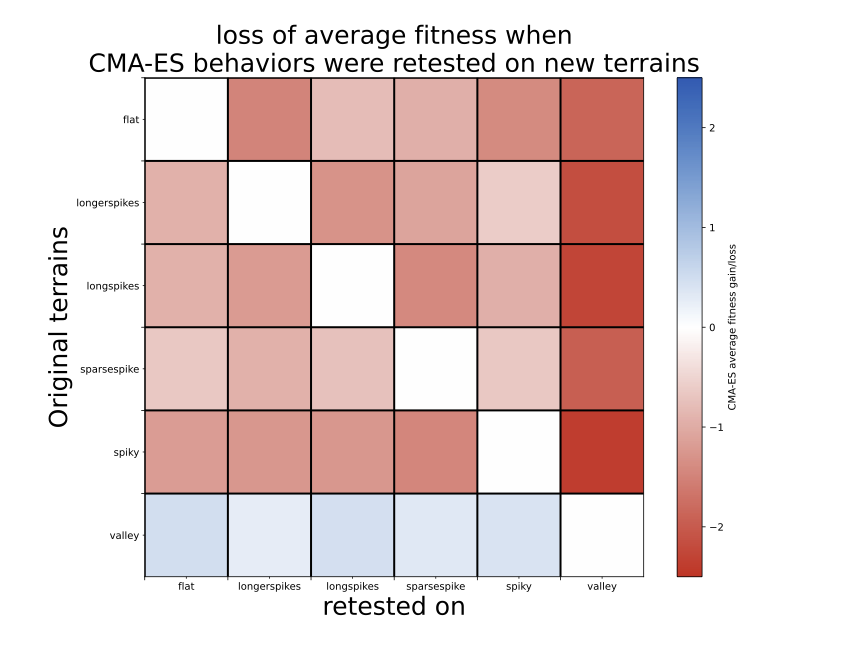}
\includegraphics[width=3.4in]{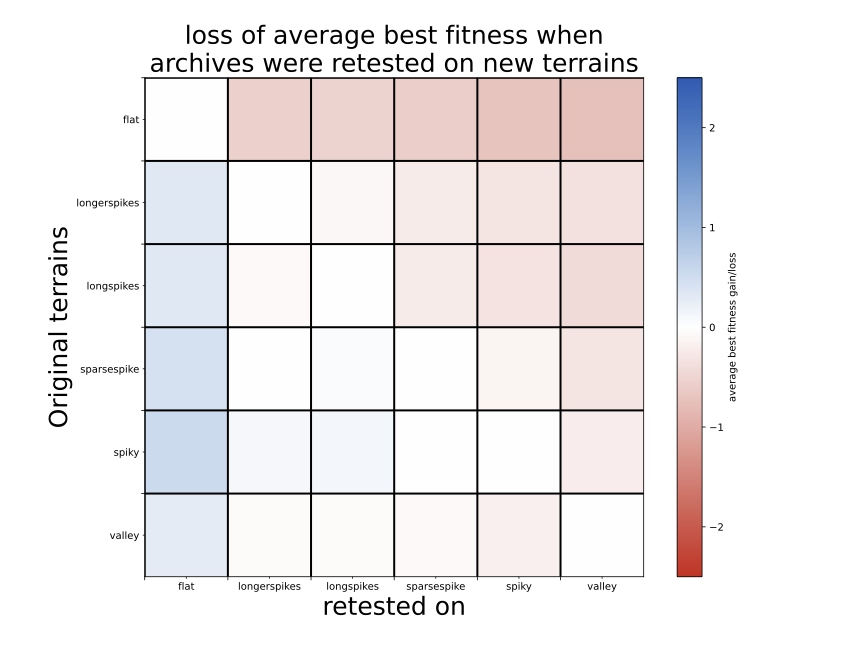}
\caption{Comparisons of absolute gain (blue) or loss (red) of fitness when gaits generated on one terrain are transferred to each other terrain.  The QDA solutions (right) lose less - and gain more - fitness when transferred than the CMA-ES solutions}
\label{fig:matrixes}
\end{figure*}

Figures~\ref{fig:boxplots} and ~\ref{fig:matrixes} illustrate the results of this experiment.  Let us begin with Figure~\ref{fig:boxplots}.  The left hand column of each subfigure shows the maximum fitness achieved by each algorithm on the initial terrain.  Not surprisingly, in every case CMA-ES substantially outperforms the Quality Diversity Algorithm.   This is intuitive, because the CMA-ES is able to focus its entire evaluation budget myopically on optimizing fitness across that one terrain, whereas the QDA has to spend its evaluation budget creating an entire archive of fit but behaviorally distinct solutions.

The remaining columns in each subfigure demonstrate the loss/gain of fitness when the solutions from the optimization phase are transferred to each of the novel terrains.   Across almost all treatments, the CMA-ES fitness drops substantially in the novel terrain, illustrating the brittle and non-robust nature of these evolved solutions.  The exception in this case is the ``valley'' terrain, which is substantially more difficult than all the other terrains - meaning that the CMA-ES solutiosn improve on the other terrains. By contrast, the best fitness from the QDA elite archive varies substantially less - and in many cases {\em increases} on the novel terrain.  There are notably increases in QDA elite fitness on new terrains -  most  often when QDA elites evolved on a non-flat terrain are transferred to the ``flat'' terrain. 

Figure~\ref{fig:matrixes} shows another perspective on these same data, aggregating the fitness gain/loss for each initial terrain.  For CMA-ES (left) we can see that, with the exception of those evolved on the ``valley'' terrain all evolved solutions are not robust when transferred to a novel terrain.  Again, however for the QDA archives (right), the fitness losses are less extreme.  In fact, unlike for the CMA-ES solutions, all of the QDA archives perform {\em better} when transferred to the flat terrain.  More interestingly the ``spiky'' terrain seems to generate archives that are more robust to novel terrains than the other terrains.

\section{Discussion and Conclusion}

In this work we've compared the robustness soft robotic gaits evolved with a single-objective optimizer, CMA-ES, against the archives of gaits generated by a Quality Diversity Algorithm.   When given the same budget of gait evaluations CMA-ES significantly outperforms QDAs in each of the terrains. 

However when those CMA-ES solutions are transferred to novel terrains their fitness drops significantly.  This in of itself is not a surprising result: CMA-ES solutions are specialized to work on the terrain they were trained on.  The archives of diverse solutions generated by the Quality Diversity Algorithm, by contrast, are more robust to transfer.  This does not mean that the archive elite that was best on the original terrain also does well on the novel terrain - instead it means one qualitatively distinct gaits in the archive does comparatively well on the novel terrain.   In other words, the diversity of ``wobble'' and ``squish'' in the archive allows for enough some benefit when transferred to a new terrain.

The one exception to this pattern is the ``valley'' terrain - one that both CMA-ES and QDA approaches struggle to find effective gaits for.  Nonetheless, the CMA-ES solutions actually increase in fitness when transferred to other terrains.  The QDA solutions generated on the spiky terrain by contrast only improves on the flat terrain.

Compellingly, the ``spiky'' terrain seems to elicit the most robustness from the QDA archives, which drop in fitness only on the challenging ``valley'' terrain.  

We can draw two insights from these results.  The first is that the time spent generating diverse archives of gaits with QDAs has on one hand a cost in terms of discovering less optimal solutions than CMA-ES.  However this investment in diversity pays dividends in the long run, by providing some inherent robustness to terrain change.   This could be leveraged in future work, for instance, by using a QDA to develop an archive of behaviorally distinct gaits for a soft robot, and using a policy network to determine which gait is most appropriate when a terrain changes.   The second insight is that the choice of terrain that gaits are generated in can have a significant effect upon the ability of those solutions to robustly transfer to a new terrain.  For CMA-ES this happens to the the challenging ``valley'' terrain, and for QDA it is the spiky terrain.  In future work we hope to explore how to find terrains that can somehow maximize the robustness of a generated archive.


\section*{Acknowledgment}

This work was supported by NSF Award ID NRI-1939930, and by the Union College Summer Research Fellowhsip program.  The authors would also like to thank the ChameleonCloud project for access to reliable high-speed baremetal compute instances.



%

\bibliographystyle{IEEEtran}
\bibliography{robosoft2024}

\end{document}